\documentclass[lettersize,journal]{IEEEtran}
\usepackage{amsmath,amsfonts}
\usepackage{algorithmic}
\usepackage{algorithm}
\usepackage{array}
\usepackage[caption=false,font=normalsize,labelfont=sf,textfont=sf]{subfig}
\usepackage{textcomp}
\usepackage{stfloats}
\usepackage{url}
\usepackage{verbatim}
\usepackage{graphicx}
\usepackage{cite}

\usepackage{float}
\usepackage{booktabs}
\usepackage{multirow}
\usepackage{float}
\usepackage[table]{xcolor}
\definecolor{best}{RGB}{255,245,204}    
\definecolor{second}{RGB}{231,240,253}  
	\definecolor{bestcolor}{RGB}{255,245,204}
	\newcommand{\best}[1]{\cellcolor{bestcolor}{\rule[-0.3ex]{0pt}{2.6ex}\textbf{#1}}}

	\usepackage{caption}
	\usepackage{booktabs}
	\usepackage{siunitx}
	\usepackage[table]{xcolor}
	\usepackage{pifont} 
	\newcommand{\cmark}{\ding{51}} %
	\newcommand{\xmark}{\ding{55}} %

\begin{document}

\title{Urban Neural Surface Reconstruction \\from Constrained Sparse Aerial Imagery with 3D SAR Fusion}

\author{Da Li, Chen Yao, Tong Mao, Jiacheng Bao, Houjun Sun
	\thanks{Corresponding Author: Jiacheng Bao}
	\thanks{da\_li@bit.edu.cn, baojiacheng@bit.edu.cn}}


\markboth{XXXX,~Vol.~x, No.~x, x~2026}%
{Shell \MakeLowercase{\textit{et al.}}: A Sample Article Using IEEEtran.cls for IEEE Journals}


\maketitle

\begin{figure*}
	\centering
	\includegraphics[width=1\linewidth]{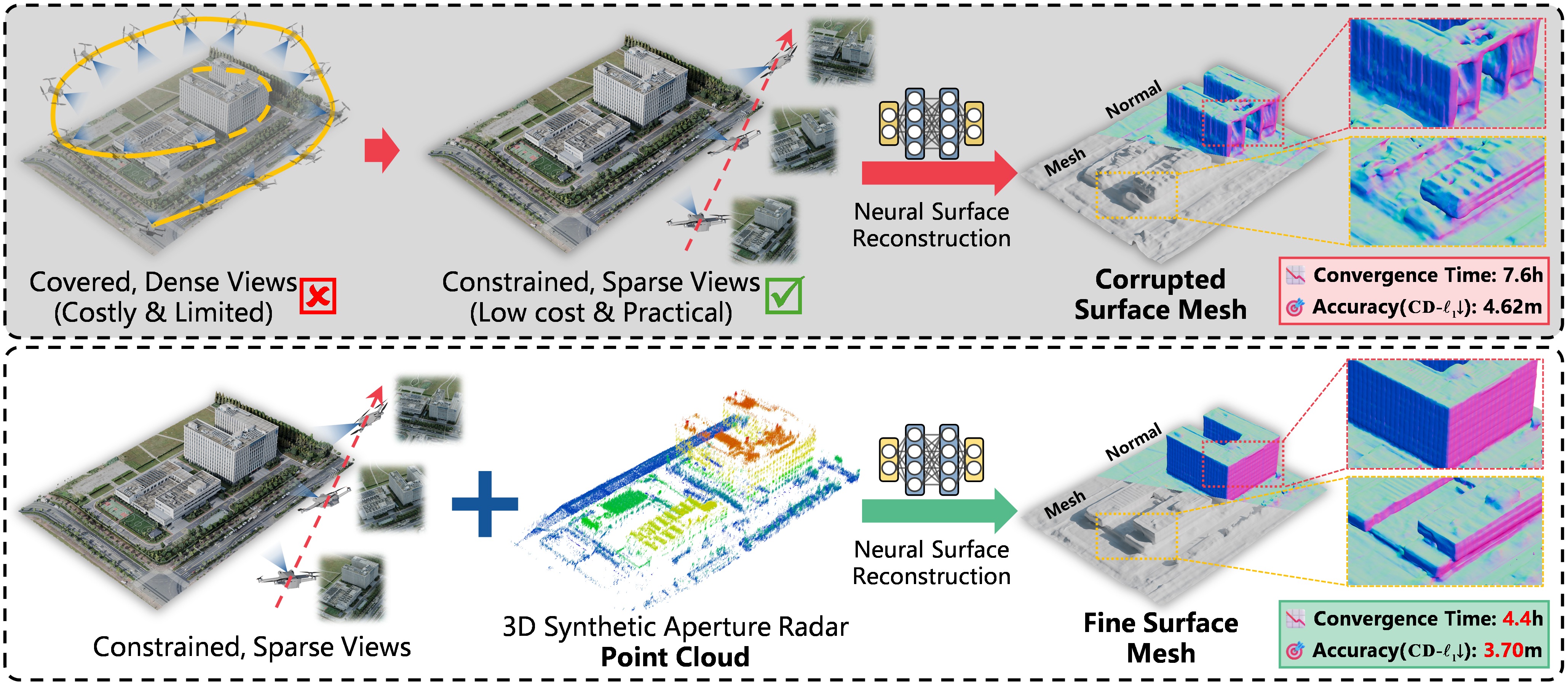}
	\caption{Overview of proposed method. Fusing constrained sparse-view aerial imagery with 3D SAR point clouds provides strong geometric priors for neural surface reconstruction, yielding finer geometry, faster convergence, and higher reconstruction accuracy.}
	\label{fig:abs}
\end{figure*}

\begin{abstract}
	Neural surface reconstruction (NSR) has recently shown strong potential for urban 3D reconstruction from multi-view aerial imagery. However, existing NSR methods often suffer from geometric ambiguity and instability, particularly under sparse-view conditions. This issue is critical in large-scale urban remote sensing, where aerial image acquisition is limited by flight paths, terrain, and cost. To address this challenge, we present the first urban NSR framework that fuses 3D synthetic aperture radar (SAR) point clouds with aerial imagery for high-fidelity reconstruction under constrained, sparse-view settings. 3D SAR can efficiently capture large-scale geometry even from a single side-looking flight path, providing robust priors that complement photometric cues from images. Our framework integrates radar-derived spatial constraints into an SDF-based NSR backbone, guiding structure-aware ray selection and adaptive sampling for stable and efficient optimization. We also construct the first benchmark dataset with co-registered 3D SAR point clouds and aerial imagery, facilitating systematic evaluation of cross-modal 3D reconstruction. Extensive experiments show that incorporating 3D SAR markedly enhances reconstruction accuracy, completeness, and robustness compared with single-modality baselines under highly sparse and oblique-view conditions, highlighting a viable route toward scalable high-fidelity urban reconstruction with advanced airborne and spaceborne optical-SAR sensing.
\end{abstract}
\section{Introduction}
\label{sec:intro}

\IEEEPARstart{3}{D} digital representations of urban scenes are essential for applications such as navigation, urban planning, geographic information system (GIS) development, and virtual reality~\cite{10892091}. Consequently, urban 3D reconstruction has long been a core research topic spanning remote sensing~\cite{kuschk2013large,10887046}, photogrammetry~\cite{wu2021photogrammetry}, computer graphics~\cite{surveyurban}, and computer vision~\cite{bauchet2024simplicity}.

Among existing techniques, multi-view aerial imagery combined with multi-view stereo (MVS) has become a mainstream solution for urban 3D reconstruction~\cite{wang2024learning}. This is due to the accessibility of aerial imagery and the ability of MVS to produce detailed urban models with realistic textures and colors~\cite{jiang2021unmanned}. Recently, a new class of MVS methods—neural surface reconstruction (NSR)—has emerged~\cite{gao2022nerf,yu2022monosdf,yariv2021volume}.These approaches represent a 3D scene using neural networks that learn its signed distance function, color, and radiance fields. Scene reconstruction is then formulated as an optimization problem, where differentiable rendering enables supervision from 2D images by minimizing discrepancies between rendered and ground-truth views. This framework allows the recovery of fine-grained, high-quality 3D surfaces from 2D imagery and has shown strong potential in aerial image–based urban reconstruction.

However, NSR methods often suffer from geometric ambiguity and surface blurring~\cite{wang2024megasurf}, primarily because photometric loss alone cannot fully resolve the inherent shape–radiance ambiguity~\cite{huang2024neusurf}. The problem becomes even more pronounced under sparse-view conditions, leading to further geometric degradation. In large-scale urban reconstruction, this limitation is especially consequential because dense aerial coverage is frequently impractical under real-world terrain, weather, and flight-path constraints~\cite{surveyurban}.


Incorporating geometric priors from other sensor modalities can help alleviate the shape blurring problem in NSR methods and reduce their dependence on dense and complete observations\cite{10892091}. Common examples include LiDAR~\cite{shi2025accurate,guo2023streetsurf} and depth cameras~\cite{azinovic2022neural,yu2022monosdf}. However, these sensors remain constrained by weather sensitivity and limited detection range in large-scale urban observation.

In urban remote sensing, three-dimensional synthetic aperture radar (3D SAR) is also an important sensing modality~\cite{6832819}. Typically mounted on UAVs, aircraft, or satellites, 3D SAR captures urban scenes by transmitting and receiving electromagnetic waves during platform motion. Compared with optical or LiDAR sensors, 3D SAR offers unique advantages: it operates under all-weather, day-and-night conditions and can achieve high-resolution 3D imaging over large areas within a single flight\cite{ding2019,qiu2024}. In terms of imaging results, 3D SAR reconstructs the 3D distribution of electromagnetic scattering intensity within the observed scene, which is commonly represented as a point cloud. Each point corresponds to the spatial location of a structure with strong scattering characteristics. In urban observation, building facades, edges, and window recesses typically exhibit strong scattering responses due to the underlying electromagnetic mechanisms\cite{gernhardt2009terrasar,ferretti2002permanent}. Consequently, 3D SAR point clouds are particularly effective at capturing building edges and geometric boundaries, making them highly suitable as geometric priors to assist urban 3D reconstruction, especially in constraint flight path.

Motivated by these insights, this paper proposes a neural surface reconstruction framework that fuses aerial imagery with 3D SAR point clouds. Our approach reconstructs high-fidelity urban surfaces using highly sparse-view aerial images from a single side-looking flight path, together with 3D SAR point clouds. 
Built upon an SDF-based NSR backbone, the radar point cloud serves as a surface-position constraint, integrated with aerial imagery to guide accurate surface learning.
Furthermore, we leverage the spatial information provided by the 3D SAR to optimize the ray selection and sampling strategy, improving reconstruction accuracy, efficiency and stability.
To validate our approach, we construct the first urban reconstruction dataset containing co-registered aerial imagery and 3D SAR point clouds.
Extensive experiments demonstrate that incorporating radar priors significantly enhances reconstruction accuracy, efficiency, and robustness compared with single-modality baselines.

The main contributions of this work are as follows:
\begin{itemize}
	\item \textbf{First Urban NSR Framework Fusing 3D SAR.} We present the first NSR framework that fuses 3D SAR point clouds and aerial imagery for high-precision 3D reconstruction of urban scenes. It achieves accurate reconstruction using only sparse observations from a single, side-looking flight path, reducing both flight-planning complexity and data acquisition cost under constrained conditions.
	\item \textbf{New Benchmark Dataset Combining 3D SAR and Aerial Imagery. }We construct the first multi-view urban reconstruction dataset that combines aerial imagery with 3D SAR point clouds, introducing 3D SAR as a new modality in urban NSR and establishing a benchmark for cross-modal 3D reconstruction research.
	\item \textbf{Radar-Derived Optimization Design.} We propose structure-aware ray selection and adaptive sampling strategies guided by radar-derived geometric cues, enabling more reliable surface optimization under sparse-view conditions.
	\item \textbf{High-Fidelity and Robust Reconstruction.} Extensive experimental results demonstrate that integrating 3D SAR point clouds substantially improves reconstruction accuracy, efficiency, and robustness compared with single-modality methods, especially under constrained and sparse-view scenarios.
\end{itemize}

\section{Related Works}
\label{sec:rw}

\noindent \textbf{3D Reconstruction from Multi-View Aerial Imagery.}
Multi-view aerial imagery has long served as a key data source for urban 3D reconstruction, typically combined with MVS algorithms~\cite{jiang2021unmanned}.
Traditional MVS methods recover 3D points by matching multi-view correspondences, e.g., semi-global matching~\cite{hirschmuller2007stereo} and PatchMatch~\cite{bleyer2011patchmatch}.
Deep learning has further advanced MVS through data-driven models such as Ada-MVS~\cite{liu2023deep} and MS-REDNet~\cite{yu2021automatic}.
However, traditional approaches rely on hand-crafted heuristics, while learning-based ones require large labeled datasets.
More recently, Neural Radiance Fields (NeRF) have shifted focus toward implicit neural representations (INRs) for detailed 3D scene modeling~\cite{lyu20243dgsr}, showing promise for aerial image–based reconstruction.

\noindent \textbf{Neural Radiance Field: }NeRF~\cite{mildenhall2021nerf} was originally proposed for novel view synthesis, which aims to generate photorealistic images of a scene from unseen viewpoints using multi-view image observations. The key idea of NeRF is to employ a neural network to learn an implicit representation of the 3D scene from known views and synthesize new ones via rendering. Scalable variants such as Mega-NeRF~\cite{turki2022mega} and Aerial-NeRF~\cite{zhang2024aerial} extend this framework to large-scale urban scenes. However, these methods primarily target view synthesis rather than explicit surface modeling. Geometry extracted from density fields often appears noisy and structurally inaccurate~\cite{lyu20243dgsr}, motivating methods that directly model 3D surfaces.

\noindent \textbf{Neural Surface Reconstruction.}
Implicit surface functions such as occupancy or signed distance fields (SDFs) enable more precise geometric definition. Methods like NeuS~\cite{wang2021neus} and VolSDF~\cite{yariv2021volume} incorporate SDFs into NeRF, retaining photorealistic rendering while improving geometry accuracy. These NSR methods achieve high-fidelity results but still suffer from geometric ambiguity under sparse-view condition~\cite{huang2024neusurf,li2023neuralangelo,fu2022geo}.

\noindent \textbf{NSR with Auxiliary Sensor Data.}
Introducing auxiliary geometric priors can alleviate ambiguity in NSR.
Neural RGB-D~\cite{azinovic2022neural} and MonoSDF~\cite{yu2022monosdf} incorporate depth supervision, while StreetSurf~\cite{guo2023streetsurf}, StreetRecon~\cite{shi2025accurate}, and URF~\cite{rematas2022urban} leverage LiDAR point clouds. 
These cross-modal approaches, mainly for object- or street-level scenes, show that complementary geometric data can guide implicit surface learning and enhance reconstruction with limited views. However, although effective in controlled or small-scale settings, such auxiliary sensors face significant challenges when applied to large-scale urban observation.


\noindent \textbf{3D SAR in Urban Imaging.}
3D SAR has been widely used in urban remote sensing applications~\cite{jiao_urban_2020,rambour2020interferometric}. Its imaging results are typically represented as point clouds, structurally similar to LiDAR in that each point denotes the 3D spatial location of a physical structure. However, the two modalities differ fundamentally in their sensing mechanisms: LiDAR directly measures surface reflections, whereas 3D SAR captures the spatial distribution of strong electromagnetic scatterers. Consequently, geometric features such as facades, edges, and recesses—often underrepresented in LiDAR data—produce strong radar responses and can be clearly delineated in 3D SAR~\cite{ferretti2002permanent}. Recognizing this advantage, we leverage 3D SAR point clouds as complementary geometric priors to assist urban 3D reconstruction.

\section{Preliminaries}
\label{sec:pre}
\begin{figure*}[t]
	\centering
	\includegraphics[width=1\linewidth]{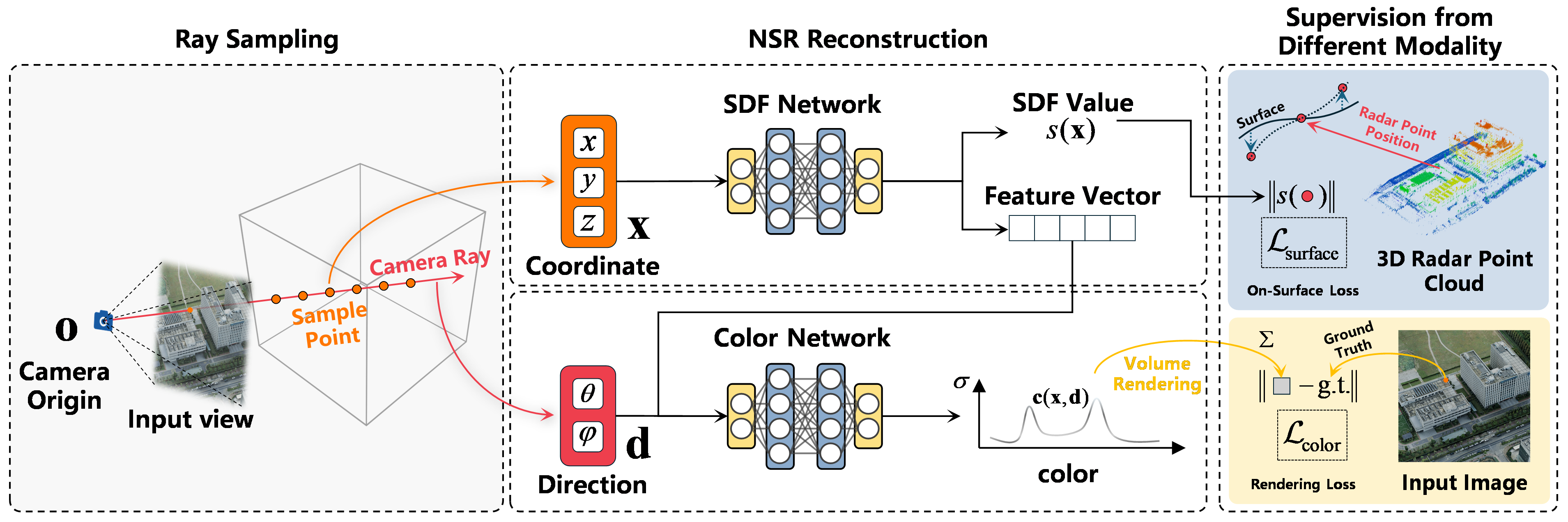}
	\caption{Proposed method framework. An SDF-based NSR network is optimized under joint supervision from aerial imagery and 3D SAR point clouds, enabling accurate and stable urban surface reconstruction.}
	\label{fig:methods}
\end{figure*}

Our method builds upon NSR frameworks that represent 3D geometry using SDFs. Since these frameworks originate from NeRF, this section first reviews NeRF and then introduces SDF-based NSR method.

\subsection{NeRF}
NeRF represents the volumetric density and color of a 3D scene using a neural network. Specifically, it employs a multilayer perceptron (MLP) to learn a mapping from a 3D spatial position $\mathbf{x}$ and a view direction $\mathbf{d}$ to the corresponding volume density $\sigma$ and color $\mathbf{c}$:
\begin{equation}
	f(\mathbf{x}, \mathbf{d})=(\sigma, c) .
\end{equation}
To render observable images for training and novel-view synthesis, NeRF adopts differentiable volume rendering. 

Given a camera ray defined by its origin $\mathbf{o}$ and direction $\mathbf{d}$, the model samples $N$ points ${\mathbf{x}_i}$ along the ray, where $\mathbf{x}_i = \mathbf{o} + t_i \mathbf{d}$ and $t_i$ is the distance from the camera center. At each sample, the network predicts a density $\sigma_i$ and color $\mathbf{c}_i$. The rendered pixel color is approximated as
\begin{equation}
	\hat{\mathbf{c}}(\mathbf{o}, \mathbf{d})=\sum_{i=1}^N w_i \mathbf{c}_i, \quad \text { where } w_i=T_i \alpha_i .
\end{equation}
Here, $\alpha_i=1-\exp \left(-\sigma_i \delta_i\right)$ denotes the opacity of the $i$-th segment, $\delta_i=t_{i+1}-t_i$ represents the distance between adjacent samples, and $T_i=\prod_{j=1}^{i-1}\left(1-\alpha_j\right)$ is the accumulated transmittance, indicating the fraction of light that reaches the camera. 

During training, the network parameters are optimized by minimizing the difference between the rendered color $\hat{\mathbf{c}}$ and the ground-truth color $\mathbf{c}$ via the color loss:
\begin{equation}
	\label{colorloss}
	\mathcal{L}_{\text {color }}=\|\mathbf{c}-\hat{\mathbf{c}}\|_1
\end{equation}
After sufficient training on multi-view images, NeRF achieves high-quality novel-view synthesis. However, because the volumetric density representation lacks explicit surface constraints, accurately recovering precise 3D geometry remains difficult.

\subsection{SDF-Based Neural Surface Reconstruction}
A SDF is a continuous field that encodes the distance from any point in space to the nearest surface, with the sign indicating whether the point lies inside (negative) or outside (positive) the surface.
Thus, a surface $\mathcal{S}$ can be implicitly defined as the zero-level set of the SDF:
\begin{equation}
	\mathcal{S}=\left\{\mathbf{x} \in \mathbb{R}^3 \mid s(\mathbf{x})=0\right\},
\end{equation}
where $s(\mathbf{x})$ denotes the signed distance at location $\mathbf{x}$.

SDF-based NSR methods extend NeRF by replacing the volumetric density $\sigma$ with the SDF value $s$ predicted by the network.
The corresponding mapping function becomes
\begin{equation}
	f(\mathbf{x}, \mathbf{d}) = (s, \mathbf{c}).
\end{equation}
To enable differentiable rendering, the volume rendering formulation is modified to interpret SDF values as opacity.
In NeuS\cite{wang2021neus}, for instance, a learnable mapping from SDF to opacity is used.
Given two consecutive samples $\mathbf{x}_i$ and $\mathbf{x}_{i+1}$ with predicted SDF values $s(\mathbf{x}_i)$ and $s(\mathbf{x}_{i+1})$, the opacity $\alpha_i$ is computed as
\begin{equation}
	\alpha_i = \max \left(
	\frac{\Phi_s(s(\mathbf{x}_i)) - \Phi_s(s(\mathbf{x}_{i+1}))}
	{\Phi_s(s(\mathbf{x}_i))}, 0 \right),
\end{equation}
where $\Phi_s(\cdot)$ denotes a sigmoid function.

This reformulation enables differentiable rendering based on SDF representations and allows the use of the same color reconstruction loss in Eq.~\ref{colorloss}.

\section{Method}
\label{sec:method}

The overall framework is shown in Fig.~\ref{fig:methods}. We adopt an SDF-based NSR network as the backbone, using both aerial imagery and 3D SAR point clouds as supervision to guide network optimization for accurate urban surface reconstruction. Leveraging radar-derived geometric priors, we further refine ray selection and sampling strategies to enhance reconstruction efficiency and stability. The following subsections describe each component.
\subsection{On-Surface Supervision from 3D SAR Point Clouds}
3D SAR point clouds capture the spatial locations of structures with strong electromagnetic scattering characteristics, such as building facades, edges, and window recesses.
3D SAR point clouds capture the locations of structures with strong electromagnetic scattering—such as facades, edges, and window recesses—typically distributed near real surfaces. Thus, radar points are assumed to lie close to the true surface, meaning their SDF values should approach zero. We therefore employ the radar point cloud as an on-surface supervision source during training. Let $\mathbf{x}_i$ denote a radar point and $s(\mathbf{x}_i)$ the predicted SDF value at that location. The surface loss is defined as:
\begin{equation}
	\mathcal{L}_{\text {surface }}=\frac{1}{N_{\text {radar }}} \sum_{i=1}^{N_{\text {radar }}}\left|\hat{S}\left(\mathbf{x}_i\right)\right|,
\end{equation}
This term enforces near-zero SDF values at radar point locations, promoting geometric accuracy.
During training, it is jointly optimized with the color loss defined in Eq.~\ref{colorloss}.

\subsection{Training Loss}
In addition to the color loss $\mathcal{L}\text{color}$ and surface loss $\mathcal{L}\text{surface}$, we apply the Eikonal loss $\mathcal{L}\text{eikonal}$—commonly used in SDF-based NSR methods—to regularize the gradient magnitude of the SDF field:
\begin{equation}
	\mathcal{L}_{\text {eikonal }}=\frac{1}{N} \sum_{i=1}^N\left(\left\|\nabla s\left(\mathbf{x}_i\right)\right\|_2-1\right)^2,
\end{equation}
where $N$ denotes the total number of sampled points along all camera rays. The overall training objective is
\begin{equation}
	\mathcal{L}_{\text{total}}=\mathcal{L}_{\text {color }}+\beta_{\text {surface}} \mathcal{L}_{\text {surface }}+\beta_{\text {eik}} \mathcal{L}_{\text {eikonal }} ,
\end{equation}
where $\beta$ terms serve as loss weights.
\begin{figure}[t]
	\centering
	\includegraphics[width=1\linewidth]{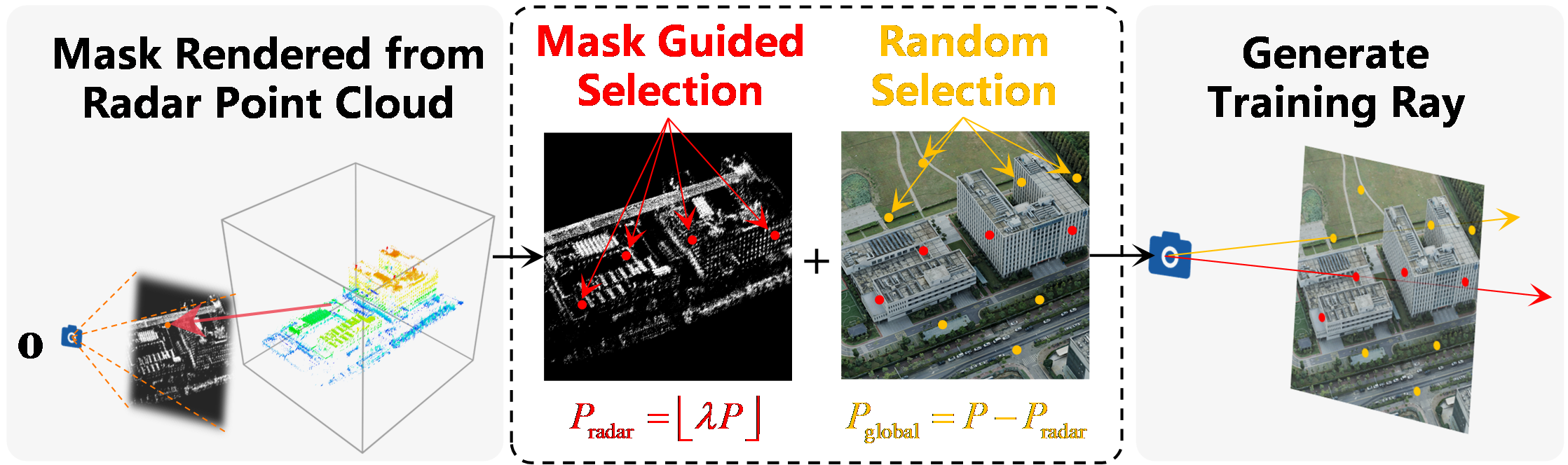}
	\caption{Structure-aware ray selection strategy.}
	\label{fig:sampling}
\end{figure}
\begin{figure}
	\centering
	\includegraphics[width=0.7\linewidth]{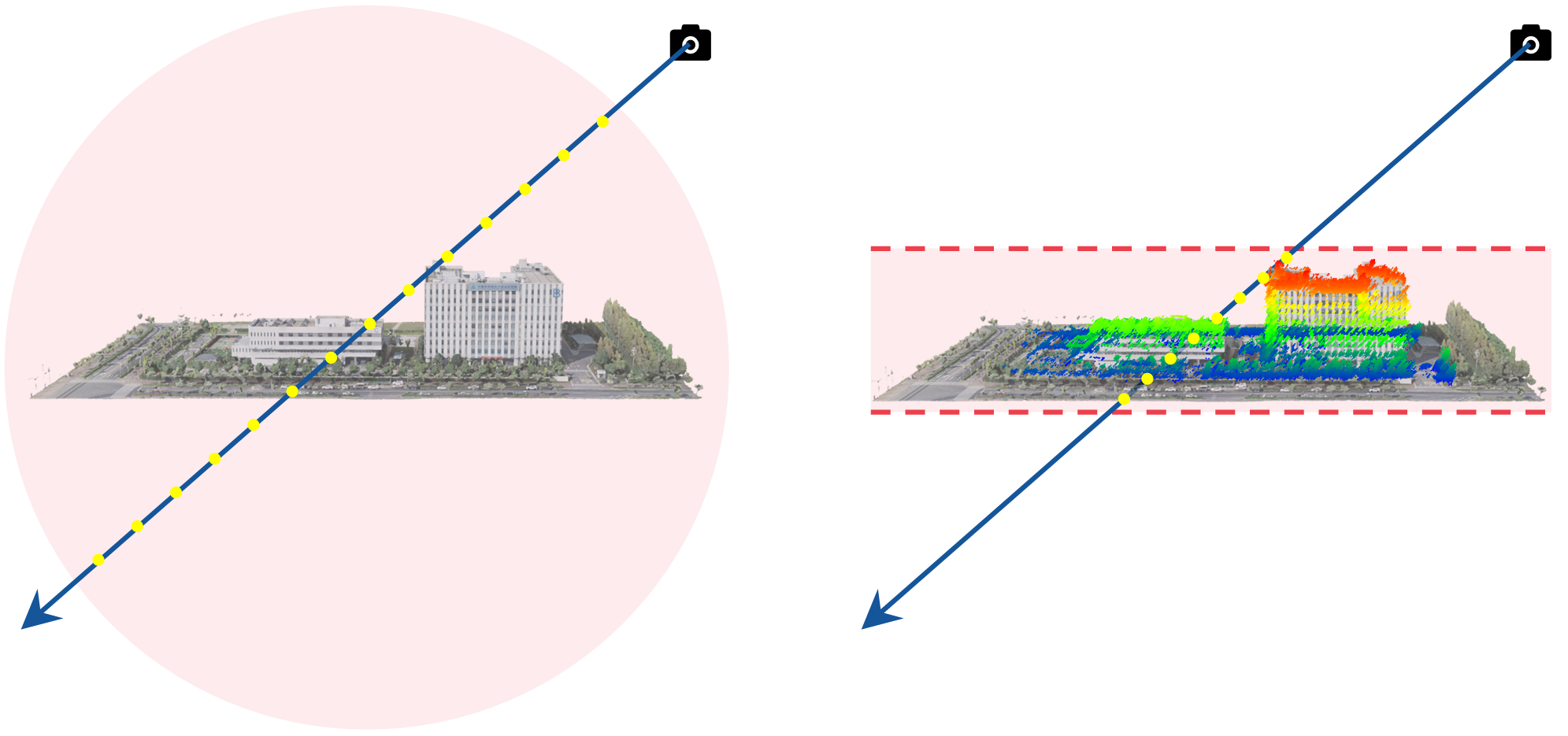}
	\caption{Geometry-constrained ray bounding strategy.}
	\label{fig:bouding}
\end{figure}
\subsection{Structure-Aware Ray Selection Guided by Radar Point Clouds}
In volumetric rendering–based training, camera rays are typically sampled randomly according to camera poses and intrinsics.
However, for large-scale urban scenes, purely random sampling often leads to uneven coverage and redundant sampling in texture-rich but geometrically uninformative regions~\cite{roessle2022dense,zhang2024aerial}.
This wastes computation and weakens structural learning. 

3D SAR point clouds, by contrast, highlight strong-scattering regions such as facades and corners, providing explicit structural cues. We therefore propose a structure-aware ray selection strategy that prioritizes radar-indicated regions.

As shown in Fig.~\ref{fig:sampling}, radar points are projected onto training images to produce a mask, where each pixel corresponds to a radar-detected structure.
During sampling, rays are drawn separately from the masked and global regions: 
\begin{equation}
	P_{\text {radar }}=\lfloor\lambda P\rfloor, \quad P_{\text {global }}=P-P_{\text {radar }},
\end{equation}

where $P$ is the total number of rays per iteration and $\lambda \in [0, 1]$ controls the sampling ratio. In our experiments, $\lambda = 0.4$ effectively balances radar-guided and global coverage, improving both efficiency and reconstruction quality.

\subsection{Geometry-Constrained Ray Bounding from Radar Point Clouds}
In addition to the ray selection strategy described above, we further exploit the geometric information provided by the 3D SAR to refine the scene sampling boundaries. Conventional methods typically normalize the scene into a fixed unit sphere or cube and perform ray sampling within this predefined volume. However, such a strategy often leads to redundant or invalid samples—particularly in urban reconstruction—where many rays may fall into empty foreground regions or underground areas. 
To address this, we estimate the ground and elevation limits from the radar point cloud and dynamically adjust the ray’s upper and lower bounds.
This adaptive boundary refinement effectively reduces redundant sampling and invalid queries, thereby improving both the efficiency and stability of the reconstruction process.

\section{Dataset Construction}
\begin{figure}[t]
	\centering
	\includegraphics[width=0.9\linewidth]{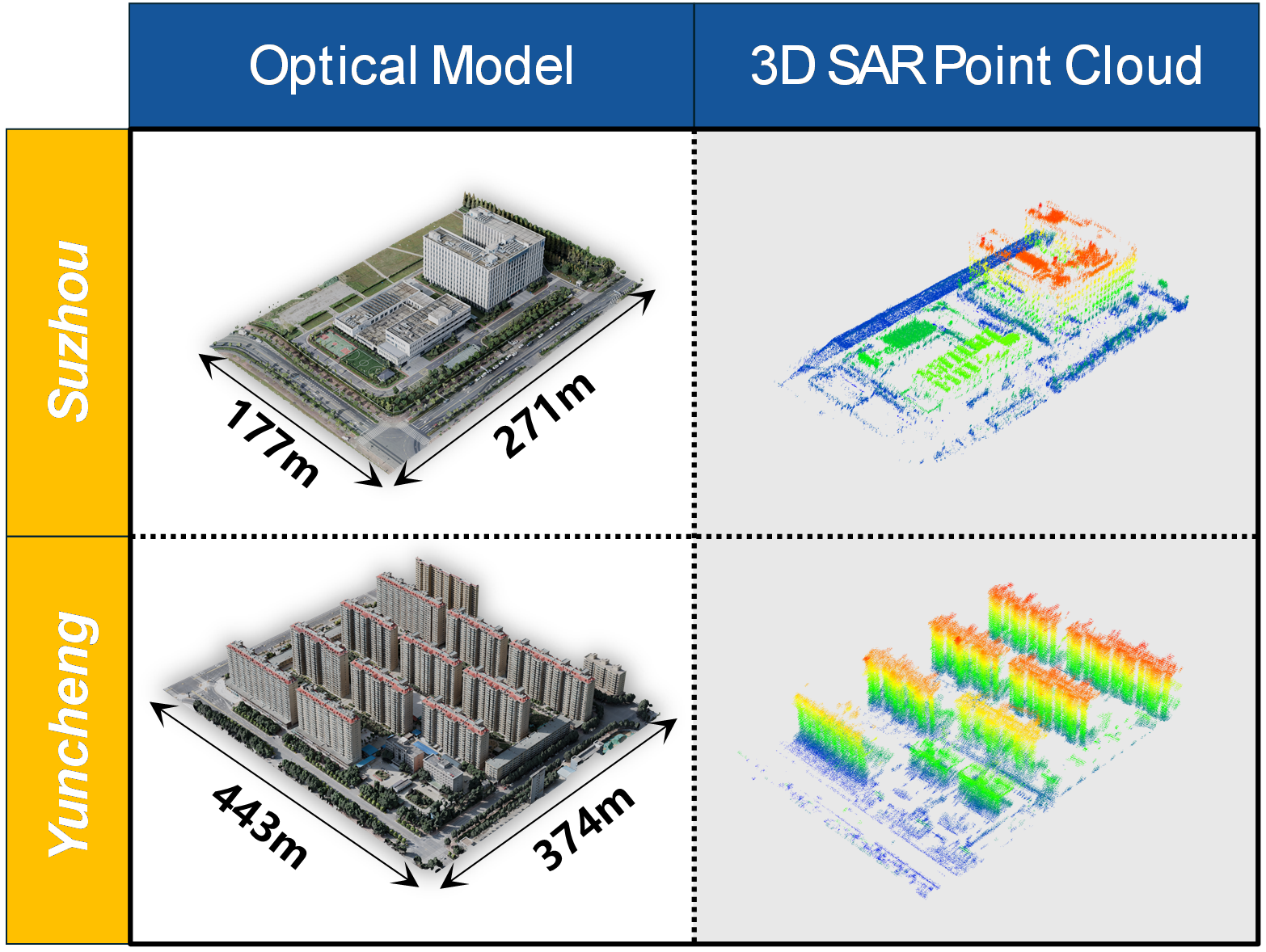}
	\caption{Examples from the SARMV3D dataset.}
	\label{fig:dataset}
\end{figure}

We construct a dataset comprising multi-view aerial images, precisely registered 3D SAR point clouds, and corresponding camera parameters for urban scenes.
To the best of our knowledge, this is the first multi-view 3D reconstruction benchmark dataset for urban environments that incorporates 3D SAR data.
The dataset is built upon the SARMV3D dataset released by the Aerospace Information Research Institute, Chinese Academy of Sciences\cite{R21112}, and further processed through image rendering, point-cloud registration, and geometric alignment. It supports research on fusing multi-view aerial imagery and 3D SAR data for urban reconstruction.
(The dataset will be publicly released at [link omitted for anonymity].)

\noindent \textbf{Overview of the SARMV3D Dataset}: SARMV3D provides SAR-based 3D imaging data, optical photogrammetric models, and high-precision LiDAR point clouds for four urban scenes, accompanied by detailed imaging parameters and flight trajectories. Among them, \textit{Suzhou} and \textit{Yuncheng} scenes contain structurally complete optical models and are selected for this study. Representative samples of their optical renderings and 3D SAR point clouds are shown in Fig.~\ref{fig:dataset}. More dataset details can be found in supplementary material.

\begin{figure}
	\centering
	\includegraphics[width=1\linewidth]{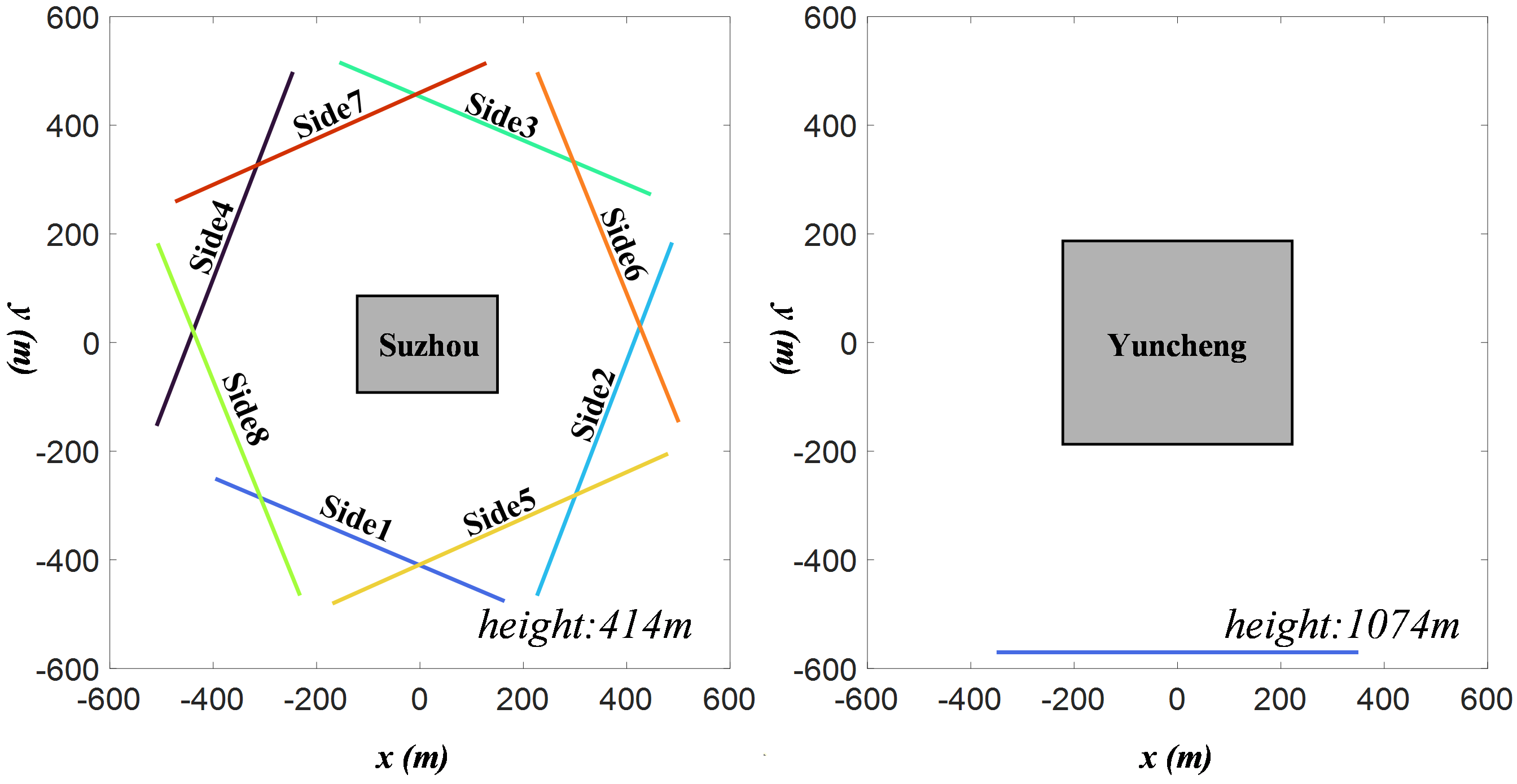}
	\caption{Flight trajectories of \textit{Suzhou} and \textit{Yuncheng} scenes.}
	\label{fig:paths}
\end{figure}

\noindent \textbf{Multi-View Aerial Image Rendering along Constrained Flight Paths}: We cleaned and trimmed the optical models, relit them in Blender under natural illumination, and rendered multi-view aerial images following the recorded flight trajectories. These trajectories directly correspond to the radar platform’s paths during 3D SAR imaging, ensuring spatial consistency between the optical renders and the 3D SAR. The \textit{Suzhou} scene contains eight trajectories distributed around the main building area, while \textit{Yuncheng} includes a single side-looking path that reflects the constrained and sparse-view setting we aim to study. The trajectory distributions are visualized in Fig.~\ref{fig:paths}.


During rendering, all cameras were oriented toward the scene center, with intrinsic parameters and image resolutions determined by scene scale and flight altitude. Notably, the \textit{Yuncheng} scene spans a large residential area with multiple building clusters and was rendered at a higher resolution (7680×7680) to capture fine architectural details. In contrast, the \textit{Suzhou} scene is smaller—containing only two adjacent building blocks—and was rendered at a lower resolution (1920×1920). 
Along each trajectory, 100 photorealistic aerial images were rendered using Blender’s Cycles renderer.
Detailed camera parameters are provided in Table~\ref{tab:dataset_params}

%

\noindent \textbf{3D SAR Point Cloud Preprocessing and Registration}: For each flight trajectory, the corresponding 3D SAR point cloud was transformed and cropped to match the optical model’s geometry.
We then applied the Iterative Closest Point (ICP) algorithm for precise registration, ensuring accurate spatial alignment between the radar data and rendered images.

Ultimately, the final dataset includes eight data groups for \textit{Suzhou} and one for \textit{Yuncheng}.
Each group contains 100 rendered aerial images, the registered 3D SAR point cloud, and corresponding camera intrinsic and extrinsic parameters.

\begin{table}
	\centering
	\caption{Main parameters of the \textit{Suzhou} and \textit{Yuncheng} datasets.}
	\resizebox{0.7\linewidth}{!}{
		\begin{tabular}{lcc}
			\toprule
			\textbf{Parameter} & \textbf{\textit{Suzhou}} & \textbf{\textit{Yuncheng}} \\
			\midrule
			Scene size (m) & 177 $\times$ 271 & 443 $\times$ 374 \\
			Number of flight paths & 8 & 1 \\
			Flight path altitude (m) & 414 & 1074 \\
			Rendering resolution & 1920 $\times$ 1920 & 7680 $\times$ 7680 \\
			Field of view (FOV) & 16$^{\circ}$ & 18$^{\circ}$ \\
			\bottomrule
	\end{tabular}}
	\label{tab:dataset_params}
\end{table}

\section{Experiments}
\label{sec:exp}
\begin{figure*}[t]
	\centering
	\includegraphics[width=0.9\linewidth]{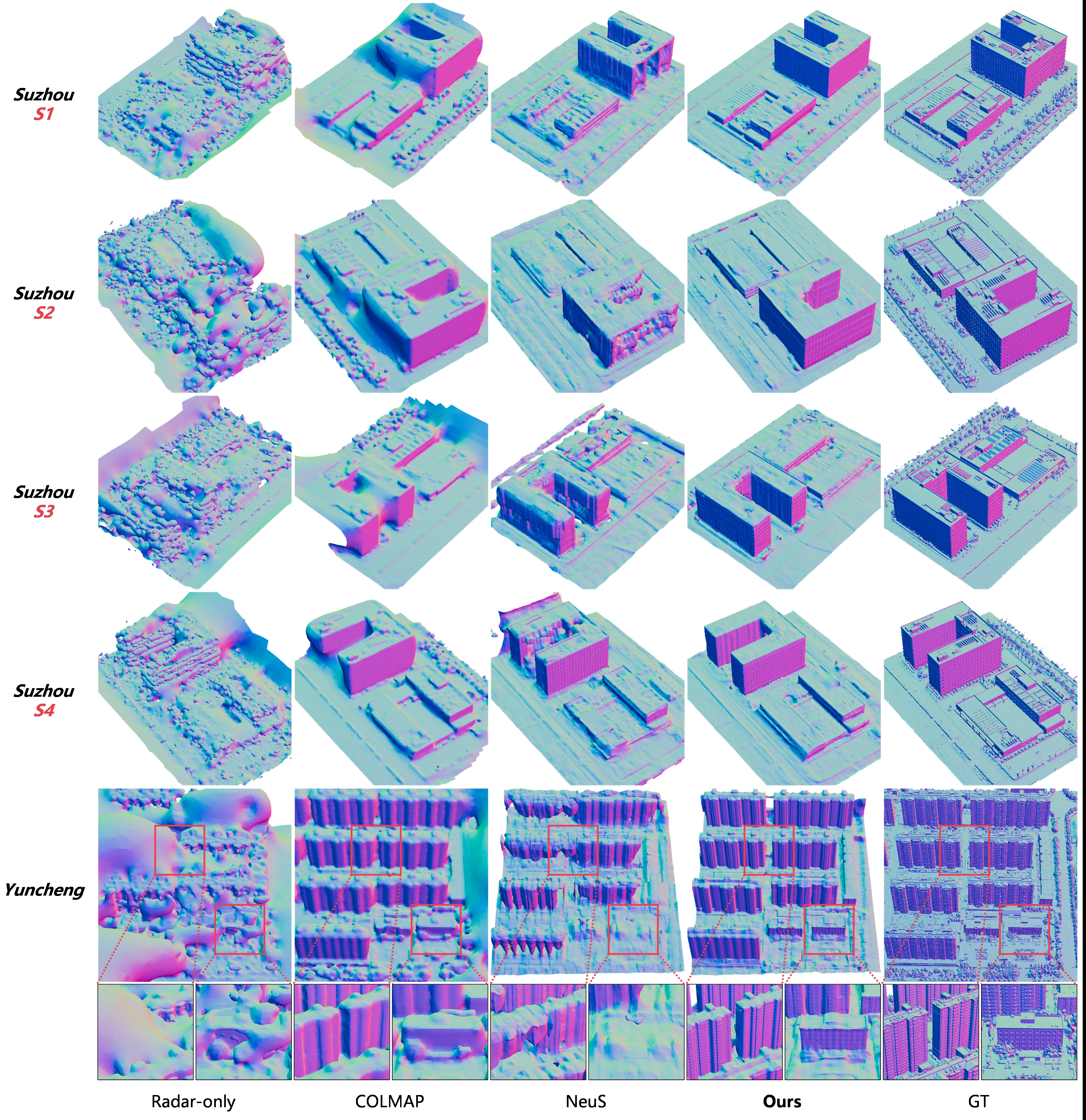}
	\caption{Visual comparison of surface reconstruction on \textit{Suzhou} and \textit{Yuncheng} scenes.}
	\label{fig:exp1}
\end{figure*}
\begin{figure*}[t]
	\centering
	\includegraphics[width=0.618\linewidth]{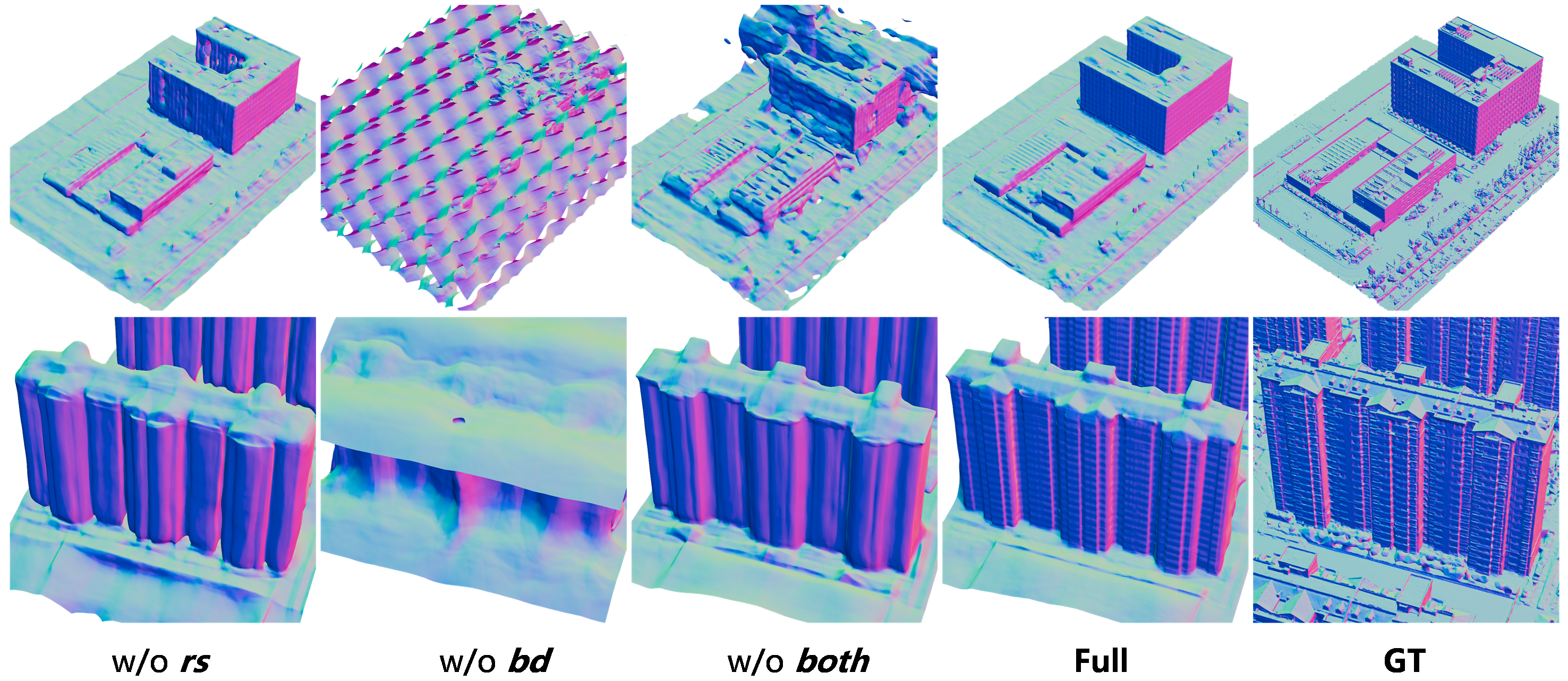}
	\caption{Ablation results on \textit{Suzhou}-S1 and \textit{Yuncheng} scenes.}
	\label{fig:exp2}
\end{figure*}
\begin{table*}[t]
	\centering
	\caption{Quantitative comparison of different reconstruction methods on the \textit{Suzhou} and \textit{Yuncheng} scenes.}
	\label{tab2}
	\resizebox{\linewidth}{!}{
		\begin{tabular}{ll|cccc|cccc|cccc|cccc}
			\toprule
			\multicolumn{2}{c}{} & \multicolumn{16}{c}{\textbf{Methods}}\\
			\cmidrule(lr){3-18}
			\textbf{City} & \textbf{Scene} &
			\multicolumn{4}{c|}{\textbf{radar-only}} &
			\multicolumn{4}{c|}{\textbf{COLMAP}} &
			\multicolumn{4}{c|}{\textbf{NeuS}} &
			\multicolumn{4}{c}{\textbf{Ours}} \\
			\cmidrule(lr){3-6}\cmidrule(lr){7-10}\cmidrule(lr){11-14}\cmidrule(lr){15-18}
			& &
			\multicolumn{1}{c}{$\mathbf{CD}{\rm{ - }}{\ell_1}\downarrow$} &
			\multicolumn{1}{c}{\textbf{Prec.}(\%)$\uparrow$} &
			\multicolumn{1}{c}{\textbf{Recall}(\%)$\uparrow$} &
			\multicolumn{1}{c|}{\textbf{F-score}(\%)$\uparrow$} &
			\multicolumn{1}{c}{$\mathbf{CD}{\rm{ - }}{\ell_1}\downarrow$} &
			\multicolumn{1}{c}{\textbf{Prec.}(\%)$\uparrow$} &
			\multicolumn{1}{c}{\textbf{Recall}(\%)$\uparrow$} &
			\multicolumn{1}{c|}{\textbf{F-score}(\%)$\uparrow$} &
			\multicolumn{1}{c}{$\mathbf{CD}{\rm{ - }}{\ell_1}\downarrow$} &
			\multicolumn{1}{c}{\textbf{Prec.}(\%)$\uparrow$} &
			\multicolumn{1}{c}{\textbf{Recall}(\%)$\uparrow$} &
			\multicolumn{1}{c|}{\textbf{F-score}(\%)$\uparrow$} &
			\multicolumn{1}{c}{$\mathbf{CD}{\rm{ - }}{\ell_1}\downarrow$} &
			\multicolumn{1}{c}{\textbf{Prec.}(\%)$\uparrow$} &
			\multicolumn{1}{c}{\textbf{Recall}(\%)$\uparrow$} &
			\multicolumn{1}{c}{\textbf{F-score}(\%)$\uparrow$} \\
			\midrule
			
			\multirow{4}{*}{\cellcolor{white}\textbf{\textit{Suzhou}}} & S1 &
			562.84 & 21.32 & 31.57 & \multicolumn{1}{c|}{25.45} &
			932.98 & 44.18 & 52.11 & \multicolumn{1}{c|}{47.82} &
			462.23 & 44.18 & 50.37 & \multicolumn{1}{c|}{47.07} &
			\best{370.10} & \best{47.38} & \best{66.05} & \best{55.18} \\
			& S2 &
			784.95 & 21.11 & 34.06 & \multicolumn{1}{c|}{26.06} &
			810.93 & 41.74 & 48.41 & \multicolumn{1}{c|}{44.83} &
			607.65 & 35.15 & 41.81 & \multicolumn{1}{c|}{38.19} &
			\best{403.28} & \best{53.75} & \best{65.73} & \best{59.14} \\
			& S3 &
			664.91 & 21.15 & 31.47 & \multicolumn{1}{c|}{25.30} &
			1045.24 & 36.81 & 43.61 & \multicolumn{1}{c|}{39.92} &
			584.28 & 27.94 & 28.65 & \multicolumn{1}{c|}{28.29} &
			\best{373.78} & \best{50.79} & \best{58.18} & \best{54.23} \\
			& S4 &
			732.13 & 20.85 & 26.81 & \multicolumn{1}{c|}{23.46} &
			601.06 & \best{54.24} & 56.33 & \multicolumn{1}{c|}{\best{55.27}} &
			580.16 & 43.06 & 55.29 & \multicolumn{1}{c|}{48.41} &
			\best{411.96} & 45.51 & \best{61.75} & 52.40 \\
			\midrule
			\textbf{\textit{Yuncheng}} & -- &
			1007.01 & 12.85 & 15.74 & \multicolumn{1}{c|}{14.14} &
			1174.73 & 30.57 & 33.63 & \multicolumn{1}{c|}{32.03} &
			767.57 & 14.13 & 13.10 & \multicolumn{1}{c|}{13.59} &
			\best{565.58} & \best{32.46} & \best{37.19} & \best{34.66} \\
			\bottomrule
	\end{tabular}}
\end{table*}

\begin{figure}
	\centering
	\includegraphics[width=0.95\linewidth]{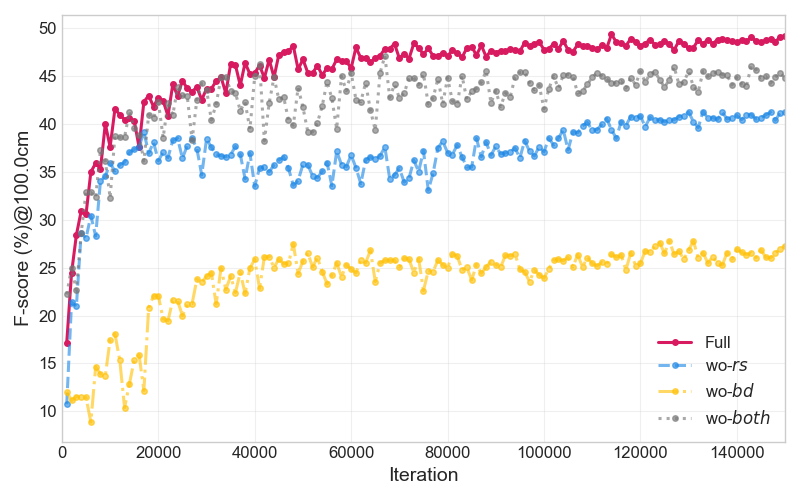}
	\caption{\textbf{F-score} convergence curves for the ablation study.}
	\label{fig:efficiency}
\end{figure}

\subsection{Experimental Setting}

\noindent \textbf{Baseline}: To evaluate the benefits of integrating 3D SAR point clouds, we compare our method against several state-of-the-art surface reconstruction approaches, including image-based methods (COLMAP and NeuS) and radar-only baselines.
For COLMAP, we initialize the pipeline from the triangulation stage using known camera parameters obtained during data acquisition. To ensure a fair comparison, Poisson surface reconstruction~\cite{kazhdan2006poisson} is applied to both COLMAP-derived and radar-derived point clouds.
Ablation studies are further conducted to assess the effects of the proposed ray selection (\textbf{\textit{rs}}) and ray bounding (\textbf{\textit{rb}}) strategies.
For the large-scale \textit{Yuncheng} scene, we adopt the scene partitioning strategy from Mega-NeRF~\cite{turki2022mega} to handle its wide spatial extent and complex geometry, which alleviates network capacity bottlenecks and improves reconstruction fidelity.


\noindent \textbf{Metrics:} Reconstruction accuracy is evaluated using standard 3D metrics, including Chamfer Distance-L1 ($\mathbf{CD}\text{-}\ell_1$), Precision (\textbf{Prec.}), \textbf{Recall}, and \textbf{F-score}. For Precision, Recall, and F-score, the distance threshold is set to 100cm.

\subsection{Results}
\begin{figure}
	\centering
	\includegraphics[width=0.8\linewidth]{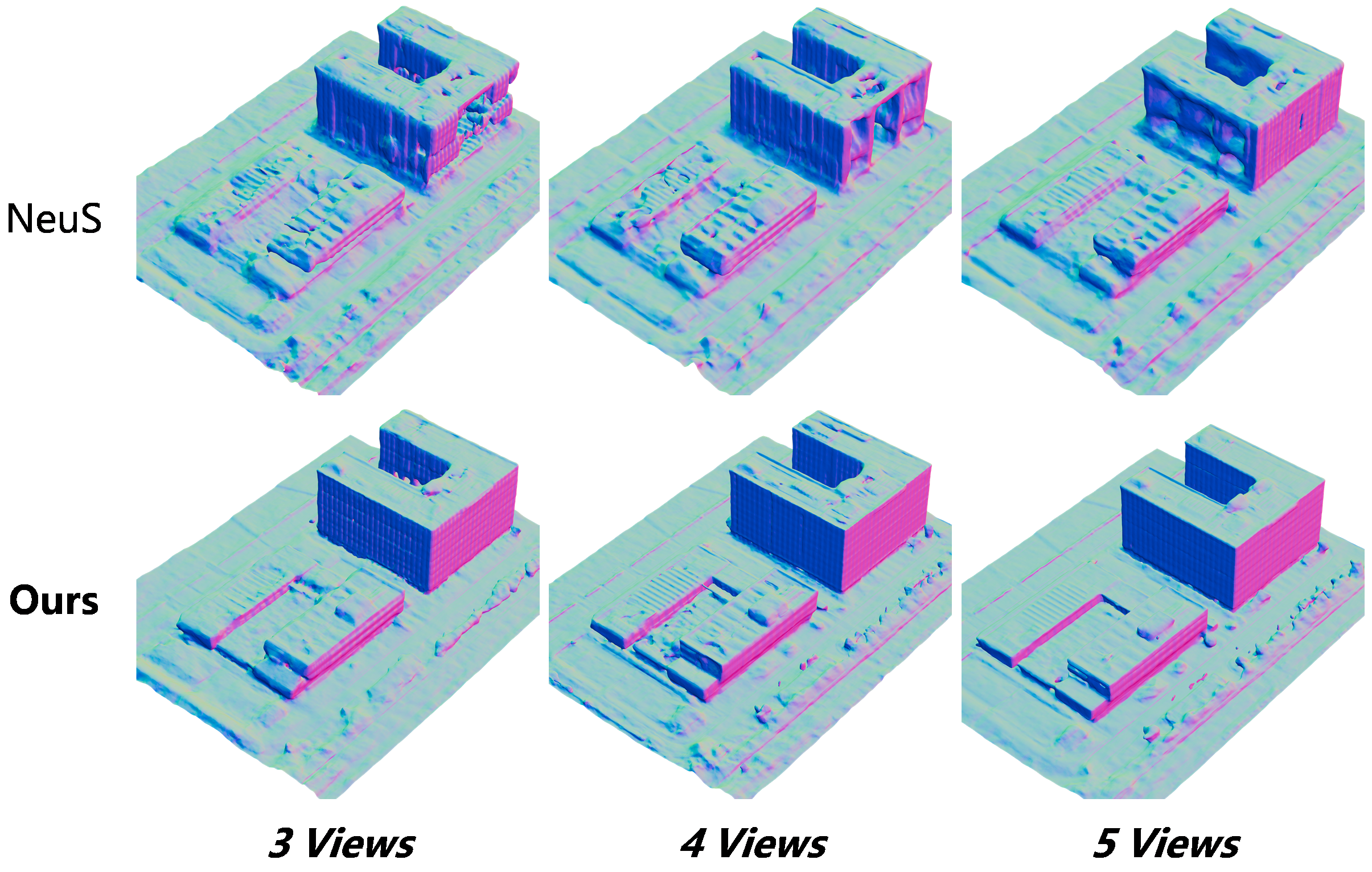}
	\caption{Reconstruction comparison under different sparse-view settings.}
	\label{fig:viewexp}
\end{figure}

\begin{table}[t]
	\centering
	\caption{Quantitative comparison for ablation on ray selection (\textbf{\textit{rs}}) and ray bounding (\textbf{\textit{bd}}).}
	\label{tab:ablation_rs_bd}
	\sisetup{table-number-alignment=center, table-text-alignment=center}
	\rowcolors{3}{gray!3}{white}
	\resizebox{0.8\linewidth}{!}{
		\begin{tabular}{cc|S[table-format=2.2]S[table-format=2.2]|S[table-format=2.2]S[table-format=2.2]}
			\toprule
			\multicolumn{2}{c|}{\textbf{Config}} &
			\multicolumn{2}{c|}{\textbf{\textit{Suzhou}}} &
			\multicolumn{2}{c}{\textbf{\textit{Yuncheng}}} \\
			\cmidrule(lr){1-2}\cmidrule(lr){3-4}\cmidrule(lr){5-6}
			\textbf{\textit{rs}} & \textbf{\textit{bd}} &
			{\textbf{Prec.(\%)}} & {\textbf{Recall(\%)}} &
			{\textbf{Prec.(\%)}} & {\textbf{Recall(\%)}} \\
			\midrule
			\xmark & \xmark & 30.10 & 58.70 & 40.10 & 44.70 \\
			\cmark & \xmark &  7.26 & 24.10 & 21.00 & 27.60 \\
			\xmark & \cmark & \best{50.50} & 60.10 & 39.70 & 42.80 \\
			\cmark & \cmark & 47.40 & \best{66.10} & \best{46.00} & \best{52.30} \\
			\bottomrule
	\end{tabular}}
\end{table}
\noindent \textbf{Qualitative Results.}
Figure~\ref{fig:exp1} shows visual comparisons among different reconstruction methods (4 views for \textit{Suzhou} and 6 for \textit{Yuncheng}).
The radar-only baseline yields irregular and noisy surfaces, reflecting the low spatial resolution and inherent ambiguity of 3D SAR point clouds.
COLMAP recovers the overall terrain and building layout but lacks sufficient texture cues, leading to geometric ambiguities and fused boundaries between adjacent structures.
NeuS reconstructs relatively complete building outlines in the small-scale \textit{Suzhou} scenes; however, facades and edges appear distorted and partially missing.
In the large-scale \textit{Yuncheng} scene with oblique aerial viewpoints, NeuS fails to recover lower buildings and collapses in complex regions.
In contrast, our method produces sharper boundaries, smoother surfaces, and more faithful building geometries that closely match the ground-truth meshes.
Furthermore, as shown in Fig.~\ref{fig:viewexp}, we compare NeuS and our method under sparse-view settings with 3, 4, and 5 input views.
It demonstrates that our approach consistently reconstructs coherent building structures with clear facades and fine textures even under extremely sparse-view conditions, whereas NeuS produces blurred geometry and lacks structural consistency across all view settings.

\noindent \textbf{Quantitative Results.}
Table~\ref{tab2} summarizes quantitative evaluations for all \textit{Suzhou}-(S1–S4) and \textit{Yuncheng} scenes.
On the small-scale \textit{Suzhou} subsets, our method achieves the lowest $\mathbf{CD}\text{-}\ell_1$ errors—reducing them by 43.7\% on average compared with other approaches—indicating a significantly closer match to ground-truth geometry.
Our framework also yields notable improvements in \textbf{Prec.} and \textbf{Recall}, increasing them by 46.8\% and 52.1\% on average, respectively.
These gains reflect more accurate point localization and more complete surface coverage.
Consequently, the \textbf{F-score} shows an average increase of 49.5\% across all scenes.
Even on the challenging large-scale \textit{Yuncheng} dataset with side-looking and oblique perspectives, our approach reduces $\mathbf{CD}\text{-}\ell_1$ from 983.1 cm to 565.6 cm and significantly improves \textbf{Prec.} (+69.2\%), \textbf{Recall} (+78.6\%), and \textbf{F-score} (+74.0\%).

\noindent \textbf{Ablation Study.}
Figure~\ref{fig:exp2} presents ablation results on a representative \textit{Suzhou} scene and a building block from the \textit{Yuncheng} scene.
Removing the ray selection strategy (w/o \textbf{\textit{rs}}) leads to blurred facades and incomplete local structures, especially around windows, as the network loses guidance toward structurally informative regions.
Without the ray bounding strategy (w/o \textbf{\textit{bd}}), reconstruction collapses entirely in both scenes, producing invalid geometry.
When both strategies are removed (w/o \textbf{\textit{both}}), the \textit{Suzhou} scene preserves coarse building outlines but exhibits cloud-like artifacts around building tops, while the \textit{Yuncheng} scene maintains overall shape yet loses fine structural and textural details.
Quantitatively, Table~\ref{tab:ablation_rs_bd} shows that enabling both strategies yields the best performance, with \textbf{Prec.}/\textbf{Recall} improvements from 30.1\%/58.7\% to 47.4\%/66.1\% on \textit{Suzhou} and from 40.1\%/44.7\% to 46.0\%/52.3\% on \textit{Yuncheng}.
Furthermore, the F-score convergence curves in Figure~\ref{fig:efficiency} reveal that the full model not only converges faster but also reaches a substantially higher and more stable plateau, whereas the variants without \textit{rs}, \textit{bd}, or both saturate early at significantly lower F-scores.
These results together confirm that both strategies are critical for efficient optimization and high-fidelity surface reconstruction.

\section{Conclusion}
We presented a novel framework that introduces 3D SAR point clouds into neural surface reconstruction for urban scenes, enabling accurate reconstruction from constrained and sparse aerial views. Beyond improved performance, our work brings 3D SAR into the NSR domain for the first time and establishes a benchmark for cross-modal urban reconstruction. Our findings further suggest that jointly leveraging optical imagery and 3D SAR provides a practical and scalable pathway for high-fidelity urban reconstruction with advanced airborne and spaceborne optical-SAR sensing. We hope this work encourages broader exploration of radar–vision fusion for learning-based 3D reconstruction under real-world operational constraints.

{
	\small
	\bibliographystyle{IEEEtran}
	\bibliography{main}

@article{surveyurban,
	author = {Musialski, P. and Wonka, P. and Aliaga, D. G. and Wimmer, M. and van Gool, L. and Purgathofer, W.},
	title = {A Survey of Urban Reconstruction},
	journal = {Computer Graphics Forum},
	volume = {32},
	number = {6},
	pages = {146-177},
	doi = {https://doi.org/10.1111/cgf.12077},
	url = {https://onlinelibrary.wiley.com/doi/abs/10.1111/cgf.12077},
	eprint = {https://onlinelibrary.wiley.com/doi/pdf/10.1111/cgf.12077},
	year = {2013}
}

@ARTICLE{10892091,
	author={Christodoulides, Andreas and Tam, Gary K. L. and Clarke, James and Smith, Richard and Horgan, Jon and Micallef, Nicholas and Morley, Jeremy and Villamizar, Nelly and Walton, Sean},
	journal={IEEE Transactions on Visualization and Computer Graphics}, 
	title={Survey on 3D Reconstruction Techniques: Large-Scale Urban City Reconstruction and Requirements}, 
	year={2025},
	volume={31},
	number={10},
	pages={9343-9367},
	doi={10.1109/TVCG.2025.3540669}}

@article{kuschk2013large,
	title={Large scale urban reconstruction from remote sensing imagery},
	author={Kuschk, Georg},
	journal={The International Archives of the Photogrammetry, Remote Sensing and Spatial Information Sciences},
	volume={40},
	pages={139--146},
	year={2013},
	publisher={Copernicus GmbH}
}

@ARTICLE{10887046,
	author={Gao, Kyle and Lu, Dening and He, Hongjie and Xu, Linlin and Li, Jonathan and Gong, Zheng},
	journal={IEEE Transactions on Geoscience and Remote Sensing}, 
	title={Enhanced 3-D Urban Scene Reconstruction and Point Cloud Densification Using Gaussian Splatting and Google Earth Imagery}, 
	year={2025},
	volume={63},
	number={},
	pages={1-14},
	doi={10.1109/TGRS.2025.3536169}}

@incollection{wu2021photogrammetry,
	title={Photogrammetry for 3D mapping in Urban Areas},
	author={Wu, Bo},
	booktitle={Urban informatics},
	pages={401--413},
	year={2021},
	publisher={Springer}
}

@inproceedings{bauchet2024simplicity,
	title={Simplicity: Reconstructing buildings with simple regularized 3d models},
	author={Bauchet, Jean-Philippe and Sulzer, Raphael and Lafarge, Florent and Tarabalka, Yuliya},
	booktitle={Proceedings of the IEEE/CVF Conference on Computer Vision and Pattern Recognition},
	pages={7616--7626},
	year={2024}
}

@article{wang2024learning,
	title={Learning-based multi-view stereo: A survey},
	author={Wang, Fangjinhua and Zhu, Qingtian and Chang, Di and Gao, Quankai and Han, Junlin and Zhang, Tong and Hartley, Richard and Pollefeys, Marc},
	journal={arXiv preprint arXiv:2408.15235},
	year={2024}
}

@article{jiang2021unmanned,
	title={Unmanned Aerial Vehicle-Based Photogrammetric 3D Mapping: A survey of techniques, applications, and challenges},
	author={Jiang, San and Jiang, Wanshou and Wang, Lizhe},
	journal={IEEE Geoscience and Remote Sensing Magazine},
	volume={10},
	number={2},
	pages={135--171},
	year={2021},
	publisher={IEEE}
}

@article{gao2022nerf,
	title={Nerf: Neural radiance field in 3d vision, a comprehensive review},
	author={Gao, Kyle and Gao, Yina and He, Hongjie and Lu, Dening and Xu, Linlin and Li, Jonathan},
	journal={arXiv preprint arXiv:2210.00379},
	year={2022}
}

@article{yu2022monosdf,
	title={Monosdf: Exploring monocular geometric cues for neural implicit surface reconstruction},
	author={Yu, Zehao and Peng, Songyou and Niemeyer, Michael and Sattler, Torsten and Geiger, Andreas},
	journal={Advances in neural information processing systems},
	volume={35},
	pages={25018--25032},
	year={2022}
}

@article{yariv2021volume,
	title={Volume rendering of neural implicit surfaces},
	author={Yariv, Lior and Gu, Jiatao and Kasten, Yoni and Lipman, Yaron},
	journal={Advances in neural information processing systems},
	volume={34},
	pages={4805--4815},
	year={2021}
}

@inproceedings{wang2024megasurf,
	title={Megasurf: Scalable large scene neural surface reconstruction},
	author={Wang, Yusen and Zhou, Kaixuan and Zhang, Wenxiao and Xiao, Chunxia},
	booktitle={Proceedings of the 32nd ACM International Conference on Multimedia},
	pages={6414--6423},
	year={2024}
}

@inproceedings{huang2024neusurf,
	title={NeuSurf: On-surface priors for neural surface reconstruction from sparse input views},
	author={Huang, Han and Wu, Yulun and Zhou, Junsheng and Gao, Ge and Gu, Ming and Liu, Yu-Shen},
	booktitle={Proceedings of the AAAI conference on artificial intelligence},
	volume={38},
	number={3},
	pages={2312--2320},
	year={2024}
}

@article{guo2023streetsurf,
	title={Streetsurf: Extending multi-view implicit surface reconstruction to street views},
	author={Guo, Jianfei and Deng, Nianchen and Li, Xinyang and Bai, Yeqi and Shi, Botian and Wang, Chiyu and Ding, Chenjing and Wang, Dongliang and Li, Yikang},
	journal={arXiv preprint arXiv:2306.04988},
	year={2023}
}

@inproceedings{azinovic2022neural,
	title={Neural rgb-d surface reconstruction},
	author={Azinovi{\'c}, Dejan and Martin-Brualla, Ricardo and Goldman, Dan B and Nie{\ss}ner, Matthias and Thies, Justus},
	booktitle={Proceedings of the IEEE/CVF Conference on Computer Vision and Pattern Recognition},
	pages={6290--6301},
	year={2022}
}

@article{shi2025accurate,
	title={Accurate and complete neural implicit surface reconstruction in street scenes using images and LiDAR point clouds},
	author={Shi, Chenhui and Tang, Fulin and Wu, Yihong and Ji, Hongtu and Duan, Hongjie},
	journal={ISPRS Journal of Photogrammetry and Remote Sensing},
	volume={220},
	pages={295--306},
	year={2025},
	publisher={Elsevier}
}

@article{jiao_urban_2020,
	title = {Urban {3D} imaging using airborne {TomoSAR}: {Contextual} information-based approach in the statistical way},
	volume = {170},
	issn = {0924-2716},
	shorttitle = {Urban {3D} imaging using airborne {TomoSAR}},
	url = {https://www.sciencedirect.com/science/article/pii/S0924271620302914},
	doi = {10.1016/j.isprsjprs.2020.10.013},
	urldate = {2024-09-25},
	journal = {ISPRS Journal of Photogrammetry and Remote Sensing},
	author = {Jiao, Zekun and Ding, Chibiao and Qiu, Xiaolan and Zhou, Liangjiang and Chen, Longyong and Han, Dong and Guo, Jiayi},
	month = dec,
	year = {2020},
	pages = {127--141},
}

@ARTICLE{6832819,
	author={Zhu, Xiao Xiang and Bamler, Richard},
	journal={IEEE Signal Processing Magazine}, 
	title={Superresolving SAR Tomography for Multidimensional Imaging of Urban Areas: Compressive sensing-based TomoSAR inversion}, 
	year={2014},
	volume={31},
	number={4},
	pages={51-58},
	doi={10.1109/MSP.2014.2312098}}

@article{ding2019,
	title={Synthetic Aperture Radar Three-dimensional Imaging——From TomoSAR and Array InSAR to Microwave Vision},
	author={Ding, Chibiao and Qiu, Xiaolan and Xu, Feng and Liang, Xingdong and Jiao, Zekun and Zhang, Fubo},
	journal={Journal of Radars},
	volume={8},
	number={6},
	pages={693--709},
	year={2019},
	publisher={Journal of Radars}
}

@article{qiu2024,
	title={Microwave Vision Three-dimensional SAR Experimental System and  Full-polarimetric Data Processing Method},
	author={QIU Xiaolan and LUO Yitong and SONG Shujie and PENG Lingxiao and CHENG Yao and YAN Qiancheng and SHANGGUAN Songtao and JIAO Zekun and ZHANG Zhe and DING Chibiao},
	journal={Journal of Radars},
	volume={13},
	number={5},
	pages={941--954},
	year={2024},
	publisher={Journal of Radars}
}

@article{gernhardt2009terrasar,
	title={TerraSAR-X high resolution spotlight persistent scatterer interferometry},
	author={Gernhardt, Stefan and Adam, Nico Alexander and Eineder, Michael and Bamler, Richard},
	journal={Proceedings of Fringe 2009},
	pages={1--5},
	year={2009}
}

@article{ferretti2002permanent,
	title={Permanent scatterers in SAR interferometry},
	author={Ferretti, Alessandro and Prati, Claudio and Rocca, Fabio},
	journal={IEEE Transactions on geoscience and remote sensing},
	volume={39},
	number={1},
	pages={8--20},
	year={2002},
	publisher={IEEE}
}

@article{liu2023deep,
	title={Deep learning based multi-view stereo matching and 3D scene reconstruction from oblique aerial images},
	author={Liu, Jin and Gao, Jian and Ji, Shunping and Zeng, Chang and Zhang, Shaoyi and Gong, JianYa},
	journal={ISPRS Journal of Photogrammetry and Remote Sensing},
	volume={204},
	pages={42--60},
	year={2023},
	publisher={Elsevier}
}

@article{hirschmuller2007stereo,
	title={Stereo processing by semiglobal matching and mutual information},
	author={Hirschmuller, Heiko},
	journal={IEEE Transactions on pattern analysis and machine intelligence},
	volume={30},
	number={2},
	pages={328--341},
	year={2007},
	publisher={IEEE}
}

@inproceedings{bleyer2011patchmatch,
	title={Patchmatch stereo-stereo matching with slanted support windows.},
	author={Bleyer, Michael and Rhemann, Christoph and Rother, Carsten},
	booktitle={Bmvc},
	volume={11},
	number={2011},
	pages={1--11},
	year={2011}
}

@article{yu2021automatic,
	title={Automatic 3D building reconstruction from multi-view aerial images with deep learning},
	author={Yu, Dawen and Ji, Shunping and Liu, Jin and Wei, Shiqing},
	journal={ISPRS Journal of Photogrammetry and Remote Sensing},
	volume={171},
	pages={155--170},
	year={2021},
	publisher={Elsevier}
}

@article{lyu20243dgsr,
	title={3dgsr: Implicit surface reconstruction with 3d gaussian splatting},
	author={Lyu, Xiaoyang and Sun, Yang-Tian and Huang, Yi-Hua and Wu, Xiuzhe and Yang, Ziyi and Chen, Yilun and Pang, Jiangmiao and Qi, Xiaojuan},
	journal={ACM Transactions on Graphics (TOG)},
	volume={43},
	number={6},
	pages={1--12},
	year={2024},
	publisher={ACM New York, NY, USA}
}

@article{mildenhall2021nerf,
	title={Nerf: Representing scenes as neural radiance fields for view synthesis},
	author={Mildenhall, Ben and Srinivasan, Pratul P and Tancik, Matthew and Barron, Jonathan T and Ramamoorthi, Ravi and Ng, Ren},
	journal={Communications of the ACM},
	volume={65},
	number={1},
	pages={99--106},
	year={2021},
	publisher={ACM New York, NY, USA}
}

@inproceedings{turki2022mega,
	title={Mega-nerf: Scalable construction of large-scale nerfs for virtual fly-throughs},
	author={Turki, Haithem and Ramanan, Deva and Satyanarayanan, Mahadev},
	booktitle={Proceedings of the IEEE/CVF conference on computer vision and pattern recognition},
	pages={12922--12931},
	year={2022}
}

@article{zhang2024aerial,
	title={Aerial-nerf: adaptive spatial partitioning and sampling for large-scale aerial rendering},
	author={Zhang, Xiaohan and Qiu, Yukui and Sun, Zhenyu and Liu, Qi},
	journal={arXiv preprint arXiv:2405.06214},
	year={2024}
}

@article{wang2021neus,
	title={Neus: Learning neural implicit surfaces by volume rendering for multi-view reconstruction},
	author={Wang, Peng and Liu, Lingjie and Liu, Yuan and Theobalt, Christian and Komura, Taku and Wang, Wenping},
	journal={arXiv preprint arXiv:2106.10689},
	year={2021}
}

@inproceedings{li2023neuralangelo,
	title={Neuralangelo: High-fidelity neural surface reconstruction},
	author={Li, Zhaoshuo and M{\"u}ller, Thomas and Evans, Alex and Taylor, Russell H and Unberath, Mathias and Liu, Ming-Yu and Lin, Chen-Hsuan},
	booktitle={Proceedings of the IEEE/CVF Conference on Computer Vision and Pattern Recognition},
	pages={8456--8465},
	year={2023}
}

@article{fu2022geo,
	title={Geo-neus: Geometry-consistent neural implicit surfaces learning for multi-view reconstruction},
	author={Fu, Qiancheng and Xu, Qingshan and Ong, Yew Soon and Tao, Wenbing},
	journal={Advances in Neural Information Processing Systems},
	volume={35},
	pages={3403--3416},
	year={2022}
}

@inproceedings{rematas2022urban,
	title={Urban radiance fields},
	author={Rematas, Konstantinos and Liu, Andrew and Srinivasan, Pratul P and Barron, Jonathan T and Tagliasacchi, Andrea and Funkhouser, Thomas and Ferrari, Vittorio},
	booktitle={Proceedings of the IEEE/CVF Conference on Computer Vision and Pattern Recognition},
	pages={12932--12942},
	year={2022}
}

@article{rambour2020interferometric,
	title={From interferometric to tomographic SAR: A review of synthetic aperture radar tomography-processing techniques for scatterer unmixing in urban areas},
	author={Rambour, Clement and Budillon, Alessandra and Johnsy, Angel Caroline and Denis, Loic and Tupin, Florence and Schirinzi, Gilda},
	journal={IEEE Geoscience and Remote Sensing Magazine},
	volume={8},
	number={2},
	pages={6--29},
	year={2020},
	publisher={IEEE}
}

@inproceedings{roessle2022dense,
	title={Dense depth priors for neural radiance fields from sparse input views},
	author={Roessle, Barbara and Barron, Jonathan T and Mildenhall, Ben and Srinivasan, Pratul P and Nie{\ss}ner, Matthias},
	booktitle={Proceedings of the IEEE/CVF conference on computer vision and pattern recognition},
	pages={12892--12901},
	year={2022}
}

@articleInfo{R21112,
	title = {SARMV3D-1.0: Synthetic Aperture Radar Microwave Vision 3D Imaging Dataset},
	journal = {Journal of Radars},
	volume = {10},
	number = {R21112},
	pages = {485},
	year = {2021},
	doi = {10.12000/JR21112},
	author = {QIU Xiaolan and JIAO Zekun and PENG Lingxiao and CHEN Jiankun and GUO Jiayi and ZHOU Liangjiang and CHEN Longyong and DING Chibiao and XU Feng and DONG Qiulei and LYU Shouye}
	}

@inproceedings{kazhdan2006poisson,
	title={Poisson surface reconstruction},
	author={Kazhdan, Michael and Bolitho, Matthew and Hoppe, Hugues},
	booktitle={Proceedings of the fourth Eurographics symposium on Geometry processing},
	volume={7},
	number={4},
	year={2006}
}
}

\vfill

\end{document}